\documentclass[11pt,a4paper]{article}
\usepackage[hyperref]{acl2020}
\usepackage{fixltx2e}
\usepackage{times}
\usepackage{latexsym}
\usepackage{url}
\usepackage{natbib}
\usepackage{graphicx}
\usepackage{comment}
\graphicspath{ {images/} }
\usepackage{amssymb}
\usepackage{amsmath}
\usepackage{bm}
\usepackage{enumitem}
\usepackage{url}
\usepackage{todonotes}
\usepackage{color}
\usepackage{tabularx}
\usepackage{url}
\usepackage[utf8]{inputenc}
\usepackage{caption}
\usepackage{subcaption}
\usepackage{color}
\usepackage[bottom]{footmisc}
\usepackage{textcomp}
\usepackage{comment}
\usepackage{multirow}
\usepackage{subcaption,siunitx,booktabs}
\usepackage{subcaption}
\usepackage{graphicx}
\usepackage{textcomp , xspace}
\usepackage{float}
\usepackage{appendix}
\usepackage{color,soul}
\usepackage{multirow}
\usepackage{subcaption}
\usepackage[skip=2.8 pt]{caption}

\usepackage{siunitx,tabularx,ragged2e,booktabs,caption}
\newcolumntype{Y}{>{\RaggedRight\arraybackslash}X}
\newcolumntype{T}[1]{S[table-format=#1]}

\makeatletter
\newcommand\xleftrightarrow[2][]{%
  \ext@arrow 9999{\longleftrightarrowfill@}{#1}{#2}}
\newcommand\longleftrightarrowfill@{%
  \arrowfill@\leftarrow\relbar\rightarrow}
\makeatother

\newcommand{\AngleParenthesis}[1]{$\langle$#1$\rangle$}

\aclfinalcopy 
\def\aclpaperid{2808} 

\title{Code and Named Entity Recognition in StackOverflow}

\author{Jeniya Tabassum, Mounica Maddela, Wei Xu, Alan Ritter\\
	    Department of Computer Science and Engineering\\
	    The Ohio State University\\
	    {\tt \{tabassum.13, maddela.4, xu.1265, ritter.1492\}@osu.edu}
 }

\begin{document}



 
\maketitle

\begin{abstract}

There is an increasing interest in studying natural language and computer code together, as large corpora of programming texts become readily available on the Internet. For example, StackOverflow currently has over 15 million programming related questions written by 8.5 million users. Meanwhile, there is still a lack of fundamental NLP techniques for identifying code tokens or software-related named entities that appear within natural language sentences. In this paper, we introduce a new named entity recognition (NER) corpus for the computer programming domain, consisting of 15,372 sentences annotated with 20 fine-grained entity types. We trained in-domain BERT representations (BERTOverflow) on 152 million sentences from StackOverflow, which lead to an absolute increase of +10 F\textsubscript{1} score over off-the-shelf BERT. We also present the SoftNER model which achieves an overall 79.10 F$_1$ score for code and named entity recognition on StackOverflow data. Our SoftNER model incorporates a context-independent code token classifier with corpus-level features to improve the BERT-based tagging model.\footnote{Our code and data are available at: {\url{https://github.com/jeniyat/StackOverflowNER/}}}

\end{abstract}

\section{Introduction}\label{intro}

Recently there has been significant interest in modeling human language together with computer code \cite{quirk2015language,iyer2016summarizing,yin2018tranx}, as more data becomes available on websites such as StackOverflow and GitHub. This is an ambitious yet promising direction for scaling up language understanding to richer domains. Access to domain-specific NLP tools could help a wide range of downstream applications. For example, extracting software knowledge bases from text \cite{movshovitz2015kb}, developing better quality measurements of StackOverflow posts  \cite{ravi2014great}, finding similar questions \cite{shirani2019question} and more. However, there is a lack of NLP resources and techniques for identifying software-related named entities (e.g., variable names or application names) within natural language texts. 

 \begin{figure}[t]
     \centering
     \includegraphics[width=0.45\textwidth]{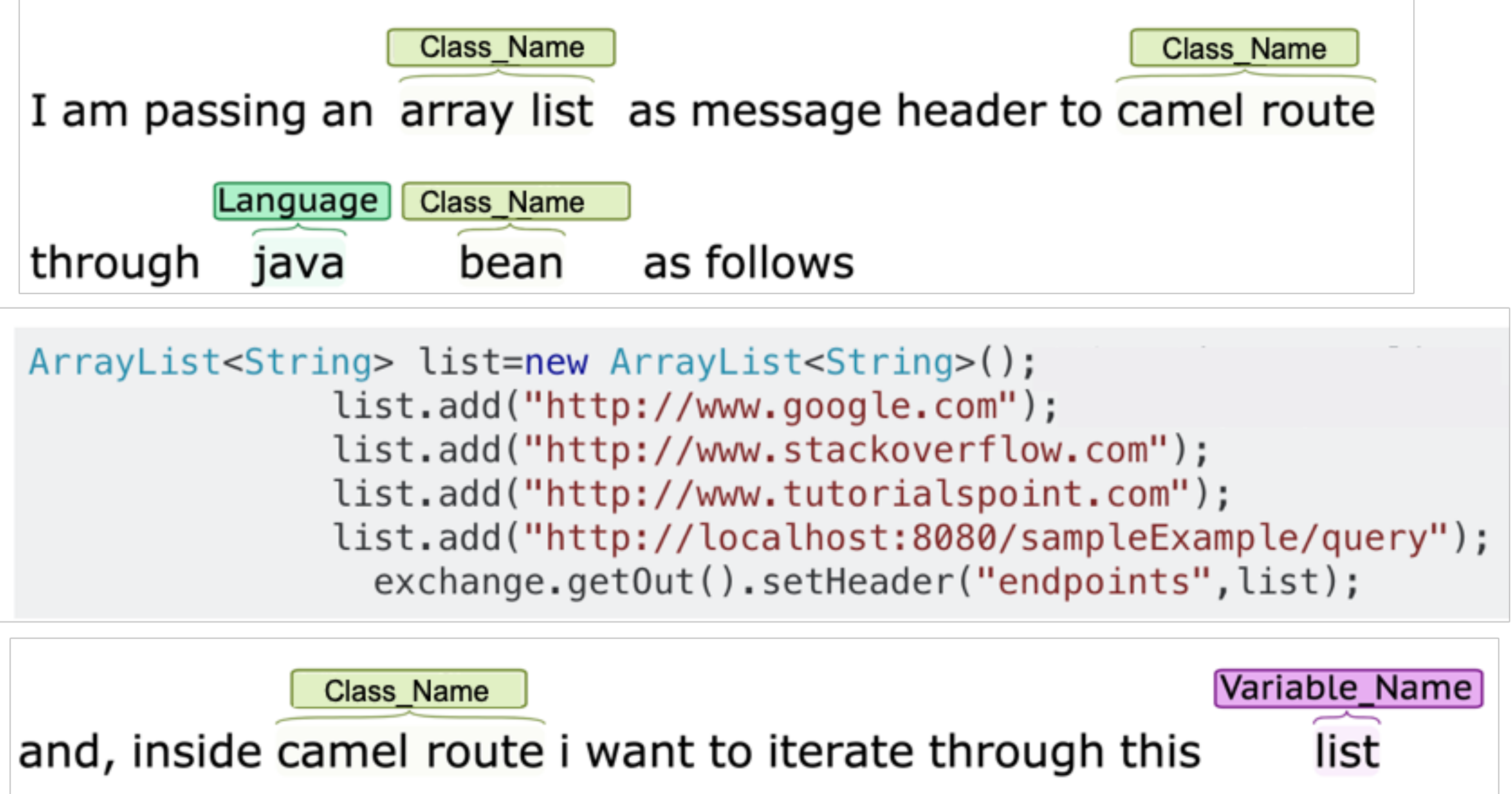}
     \caption{Examples of software-related named entities in a StackOverflow post.}
    
    \label{fig:code_block_assist}
    \vspace{-.3 cm}
 \end{figure}

In this paper, we present a comprehensive study that investigates the unique challenges of named entity recognition in the social computer programming domain.  These named entities are often ambiguous and have implicit reliance on the accompanied code snippets. For example, the word `\textit{list}' commonly refers to a data structure, but can also be used as a variable name (Figure \ref{fig:code_block_assist}). In order to recognize these entities, we propose a software-related named entity recognizer (SoftNER) that utilizes an attention network to combine the local sentence-level context with corpus-level information extracted from the code snippets. Using our newly annotated corpus of 15,372 sentences in StackOverflow, we rigorously test our proposed SoftNER model, which outperforms both a BiLSTM-CRF and fine-tuned BERT model when identifying 20 types of software-related named entities. Our key contributions are the following:

\begin{itemize}[noitemsep]
  
  \item A new StackOverflow NER corpus manually annotated with 20 entity types, including in-line code within natural language sentences (\S \ref{sec:dataset_and_annotation}).  We demonstrate that NER in the software domain is an ideal benchmark for testing the effectiveness of contextual word representations, such as ELMo \cite{Peters:2018} and BERT \cite{devlin2018bert}, due to its inherent polysemy and salient reliance on context. 

  \item An in-domain trained neural SoftNER tagger for StackOveflow (\S \ref{sec:softner}) that can recognize 20 fine-grained named entity types related to software developing. We also tested its performance on GitHub data of readme files and issue reports.  
  
  \item A code token recognizer (\S \ref{sec:domain_knowledge}) that utilizes StackOveflow code snippets to capture the spelling patterns of code-related tokens, and consistently improves the NER tagger.

  \item In-domain pretrained ELMo and BERT representations (\S \ref{sec:implementation}) on 152 million sentences from StackOverflow that significantly outperforms off-the-shelf ELMo and leads to more than 21 points increase in F\textsubscript{1} score over off-the-shelf BERT.

\end{itemize}

\noindent Overall, our named entity tagger (SoftNER) achieves a 79.10\% F\textsubscript{1} score on StackOverflow and 61.08\% F\textsubscript{1} score on GitHub data for extracting the 20 software related named entity types. We believe this performance is sufficiently strong to be practically useful. We have released our data and code, including the named entity tagger, our annotated corpus, annotation guideline, a specially designed tokenizer, and pre-trained StackOverflow BERT and ELMo embeddings.

\section{Annotated StackOverflow Corpus}\label{sec:dataset_and_annotation}

In this section, we describe the construction of our StackOverflow NER corpus. We randomly selected  1,237 question-answer threads from StackOverflow  10-year archive (from September 2008 to March 2018) and manually annotated them with 20 types of entities. For each question, four answers were annotated, including the accepted answer, the most upvoted answer, as well as two randomly selected answers (if they exist). Table \ref{tab:corpus_data_stat} shows the statistics of our corpus. 40$\%$ of the question-answer threads were double-annotated, which are used as the development and test sets in our experiments (\S \ref{sec:experiments}). We also annotated 6,501 sentences from GitHub readme files and issue reports as additional evaluation data.

\subsection{Annotation Schema}\label{annotation_scheme}

\begin{table}[t]
\begin{subtable}{.5\textwidth} 

\footnotesize
\begin{tabular}{l|rrr|r}
\hline
 & \textbf{ Train} & \textbf{ Dev}   & \textbf{Test}  & \textbf{Total} \\
\hline
\#questions  \hspace{.08 in}      & 741   & 247   & 249   & 1,237   \\
\#answers          &  897  & 289   & 315   & 1,501   \\
\hline
\#sentences        & 9,315  & 2,942  & 3,115  & 15,372  \\
\#tokens           & 136,996 & 43,296 & 45,541 & 225,833 \\
\#entities         & 11,440  & 3,949   & 3,733  & 19,122 \\
\hline
\end{tabular}
\label{tab:corpus_data_stat_total}

\end{subtable}

\vspace{1mm}

\begin{subtable}{.1\textwidth} 

\centering
\footnotesize
\begin{tabular}{l|rr}
\hline
  & \textbf{ per  Question} & \textbf{\hspace{.02 in} per Answer}  \\
\hline
avg. \#sentences     & 6.84   & 4.60    \\
avg. \#tokens         & 98.46  & 69.37 \\
avg. \#entities     & 7.62   & 5.11  \\  
avg. \#tokens per sent. & 14.38   & 15.08  \\
\hline
\end{tabular}
\label{tab:corpus_data_stat_avg}

\end{subtable}
\caption{Statistics of our StackOverflow NER corpus. These counts exclude all the code blocks and output blocks (i.e., lines that appear within \AngleParenthesis{code} and \AngleParenthesis{blockquote} tags).}
\label{tab:corpus_data_stat}
\end{table}


We defined and annotated 20 types of fine-grained entities, including 8 code-related entities and  12 natural language entities. The code entities include mentions of  {\sc Class},  {\sc Variable},  {\sc In Line Code}, {\sc Function}, {\sc Library}, {\sc Value},  {\sc Data Type}, and {\sc HTML XML Tag}. Whereas the natural language entities  include mentions of  {\sc Application}, {\sc UI Element},  {\sc Language}, {\sc Data Structure},   {\sc Algorithm}, {\sc File Type}, {\sc File Name}, {\sc Version},  {\sc Device}, {\sc OS}, {\sc Website}, and {\sc User Name}.

Our annotation guideline was developed through several pilots and further updated with notes to resolve difficult cases as the annotation progressed.\footnote{Our  annotation guideline is available at: {\url{https://github.com/jeniyat/StackOverflowNER/}}.} Each entity type was defined to encourage maximum span length (e.g., `\textit{SGML parser}' instead of `\textit{SGML}'). We annotated noun phrases without including modifiers (e.g., `\textit{C}' instead of `\textit{Plain C}'), except a few special cases (e.g., `\textit{rich text}' as a common {\sc File Type}). On average, an entity contains about 1.5 tokens.  While {\sc Variable}, {\sc Function} and {\sc Class} names mostly consist of only a single token, our annotators found that some are written as multiple tokens when mentioned in natural language text (e.g.,  `\textit{array list}' for `\textit{ArrayList}' in Figure \ref{fig:code_block_assist}). The annotators were asked to read relevant code blocks or software repositories to make a decision, if needed. Annotators also searched Google or Wikipedia to categorize unfamiliar cases.  

The annotators were asked to update, correct, or add annotations from the user provided \AngleParenthesis{code} markdown tags. StackOverflow users can utilize  \AngleParenthesis{code} markdowns to highlight the code entities within the natural language sentences. However, in reality, many users do not enclose the code snippets within the \AngleParenthesis{code} tags; and sometimes use them to highlight non-code elements, such as email addresses, user names, or natural language words. While creating the StackOverflow NER corpurs,  we found that 59.73\% of code-related entities are not marked by the StackOverflow users.  Moreover, only 75.54\% of the \AngleParenthesis{code} enclosed texts are actually code-related, while 10.12\% used to are highlighting natural language texts. The rest of cases are referring to non-code entities, such as {\sc{Software Name}s} and {\sc{Version}s}. While markdown tag could be a useful feature for entity segmentation (\S \ref{sec:segmenter}), we emphasize the importance of having a human annotated corpus for training and evaluating NLP tools in the software domain.

\begin{figure*}[tbhp]
    \centering
    \includegraphics[width=6.4 in,keepaspectratio]{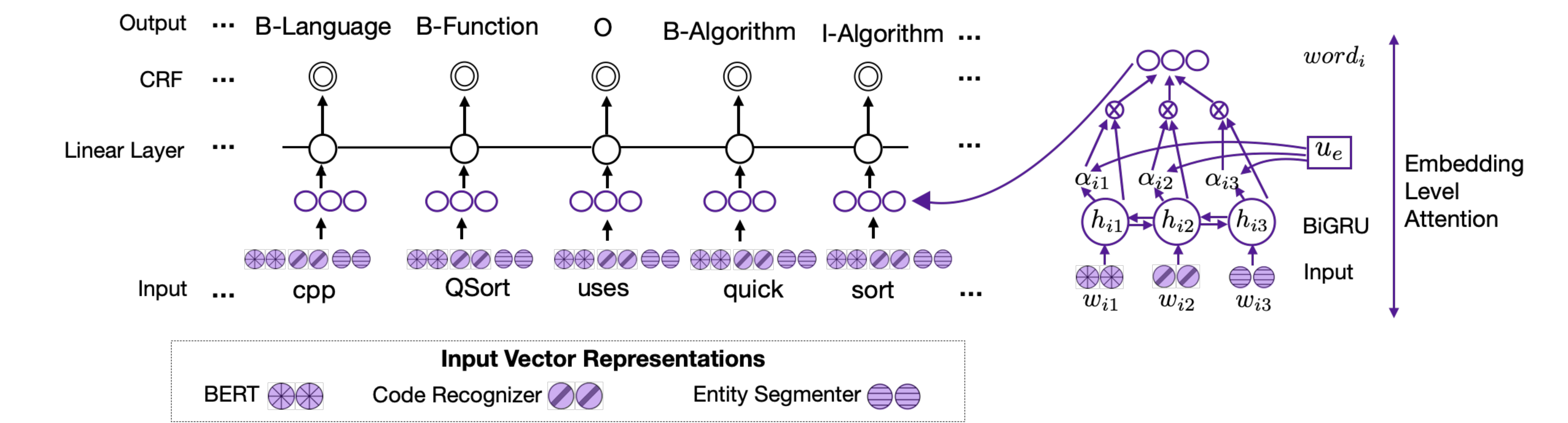}
    \vspace{.2 mm}
    \caption{Our SoftNER model. It utilizes an attention network to combine the contextual word embeddings (BERT\textsubscript{base}) with the domain-specific embeddings (Code Recognizer and Entity Segmenter). The detailed structure of the attention network is depicted on the right.}
    \label{fig:ner_model}
\end{figure*}

\subsection{Annotation Agreement}
\label{sec:annotation_process}

Our corpus was annotated by four annotators who are college students majored in computer science. We used a web-based annotation tool, BRAT \cite{stenetorp-EtAl:2012:DemoEACL2012}, and provided annotators with links to the original post on StackOverflow. For every iteration, each annotator was given 50 question-answer threads to annotate, 20 of which were double-annotated. An adjudicator then discussed disagreements with annotators, who also cross-checked the 30 single-annotated questions in each batch. The inter-annotator agreement is 0.62 before adjudication, measured by token-level Cohen’s Kappa ~\cite{cohen_kappa}.  

\subsection{Additional GitHub Data}

To better understand the domain adaptability of our work, we further annotated the readme files and issue reports from 143 randomly sampled repositories in the GitHub dump \cite{GHTorrent} (from October 29, 2007 to December 31, 2017). We removed all the code blocks from the issue reports and readme files collected from these 143 repositories. The resulting GitHub NER dataset consists of 6,510 sentences and 10,963 entities of 20 types labeled by two in-house annotators. The inter-annotator agreement of this dataset is 0.68, measured by token-level Cohen’s Kappa.

\subsection{StackOverflow/GitHub Tokenization}
\label{sec:tokenizer}

We designed a new tokenizer, \textsc{SOTokenizer}, specifically for the social computer programming domain. StackOverflow and GitHub posts exhibit common features of web texts, including abbreviations, emoticons, URLs, ungrammatical  sentences and spelling errors. We found that tokenization is non-trivial as many code-related tokens are mistakenly split by the existing web-text tokenizers, including the CMU Twokenizer \cite{Gimpel:2011:ark-twokenizer}, Stanford TweetTokenizer \cite{manning-EtAl:2014:Stanford_CoreNLP}, and NLTK Twitter Tokenizer \cite{Loper:2002:NLTK}:

\vspace{-.2 cm}
\begin{table}[H]
\hspace{-.2 cm}
\footnotesize
\begin{tabular}{p{1.1 in} p{1.7 in}}
\textit{txScope.Complete()} & [ {\textit{`txScope' `.' `Complete' `(' `)' }}] \\
\textit{std::condition\_variable} &  [ {\textit{`std' `:' `:' `condition\_variable'}}] \\
\textit{math.h} & [ {\textit{`math' `.' `h'}}] \\
\textit{\AngleParenthesis{span}} & [{\textit{`$\langle$' `span' `$\rangle$'}}]  \\
\textit{a==b} & [{\textit{`a' `=' `=' `b'}}] \\
\end{tabular}
\end{table}
\vspace{-.3 cm}

\noindent Therefore, we implemented a new tokenizer, using Twokenizer\footnote{\url{https://github.com/myleott/ark-twokenize-py}} as the starting point and added additional regular expression rules to avoid splitting code-related tokens.

 \section{Named Entity Recognition Models}
 \label{sec:softner}
 
 The extraction of software-related named entities imposes significant challenges as it requires resolving a significant amount of unseen tokens, inherent polysemy, and salient reliance on context. Unlike news or biomedical data, spelling patterns and long-distance dependencies are more crucial in the software domain to resolve ambiguities and categorize unseen words. Taken in isolation, many tokens are highly ambiguous and can  refer  to either programming concepts or common English words, such as: `\textit{go}', `\textit{react}', `\textit{spring}', `\textit{while}', `\textit{if}', `\textit{select}'. To address these challenges, we design the SoftNER model that leverages sentential context to disambiguate and domain-specific character representations to handle rare words. Figure \ref{fig:ner_model} shows the architecture of our model, which consists of primarily three components:

 \begin{itemize}[noitemsep]

  \item An \textbf{input embedding layer} (\S \ref{sec:domain_knowledge}) that extracts contextualized embeddings from the BERT\textsubscript{base} model and two new domain-specific embeddings for each word in the input sentence. 
  
  \item A \textbf{embedding attention layer} (\S \ref{sec:han}) that combines the three word embeddings using an attention network.
  
  \item A \textbf{linear-CRF} layer that predicts the entity type of each word using the attentive word representations from the previous layer.  
\end{itemize}


\subsection{Input Embeddings}
\label{sec:domain_knowledge}
  For each word in the input sentence, we extract in-domain BERT \cite{devlin2018bert} representations and two new domain-specific embeddings produced by (i) a \textbf{Code Recognizer}, which represents if a word can be part of a code entity regardless of context; and (ii) an \textbf{Entity Segmenter}, that predicts whether a word is part of any named entity in the given sentence. Each domain-specific embedding is created by passing a binary value, predicted by a network independent from the SoftNER model. We describe the two standalone auxiliary models that generate these domain-based vectors below.

\subsubsection{In-domain Word Embeddings}

Texts in the software engineering domain contain programming language tokens, such as variable names or code segments, interspersed with natural language words.  This makes input representations pre-trained on general book or Wikipedia texts unsuitable for software domain.  We pre-trained different in-domain word embeddings, including BERT (BERTOverflow), ELMo (ELMoVerflow), and GloVe (GloVerflow) vectors on the StackOverflow 10-year archive\footnote{\url{https://archive.org/details/stackexchange}} of 152 million sentences and 2.3 billion tokens (\S \ref{sec:implementation}).

\subsubsection{Context-independent Code Recognition}
\label{sec:inline_code_recognition}

Humans with prior programming knowledge can easily recognize that `\textit{list()}' is code, `\textit{list}' can be either code or a common English word, whereas `\textit{listing}' is more likely a non-code natural language token. We thus introduce a code recognition module to capture such prior probability of how likely a word can be a code token without considering any contextual information. It is worth noting that this standalone code recognition model is also useful for language-and-code research, such as retrieving code snippets based on natural language queries \cite{iyer2016summarizing, Giorgi262790, Yao:2019:CCA:3308558.3313632}

Our code recognition model (\textbf{Code Recognizer}) is a binary classifier. It utilizes language model features and spelling patterns to predict whether a word is a code entity. The input features include unigram word and 6-gram character probabilities from two language models (LMs) that are trained on the Gigaword corpus \cite{Gigaword} and all the code-snippets in the StackOverflow 10-year archive respectively. We also pre-trained FastText \cite{joulin2016fasttext} word embeddings using these code-snippets, where a word vector is represented as a sum of its character ngrams. We first transform each ngram probability into a $k$-dimensional vector using Gaussian binning \cite{EMNLP-2018-Maddela}, which has shown to improve the performance of neural models using numeric features  \cite{sil2017neural, liu2016, EMNLP-2018-Maddela}. We then feed the vectorized features into a linear layer, concatenate the output with FastText character-level embeddings, and pass them through another hidden layer with sigmoid activation. We predict the token as a code-entity if the output probability is greater than $0.5$. This binary prediction is then converted into a vector and used as an input to the SoftNER model.

\subsubsection{Entity Segmentation}\label{sec:segmenter}

The segmentation task refers to identifying entity spans without assigning entity category. Entity segmentation is simpler and less error-prone than entity recognition as it does not require a fine-grained classification of the entity types. In fact, a segmentation model (\textbf{Entity Segmenter}) trained on our annotated StackOverflow corpus can achieve 90.41\% precision on the dev set (details in \S \ref{sec:evaluation_auxiliary}), predicting whether each token is a part of entity in the given sentence. Our segmentation model fine-tunes the in-domain BERT after concatenating it with two hand-crafted features:

\begin{itemize}[noitemsep]
\item \textbf{Word Frequency}  represents the word occurrence count in the training set. As many code tokens are defined by individual users, they occur much less frequently than normal English words. In fact, code and non-code tokens have an average frequency of 1.47 and 7.41 respectively in our corpus. Moreover, ambiguous token that can be either code or non-code entities, such as `\textit{windows}', have a much higher average frequency of 92.57. To leverage this observation, we include word frequency as a feature, converting the scalar value into a $k$-dimensional vector by Gaussian binning \cite{EMNLP-2018-Maddela}. 

\item \textbf{Code Markdown} indicates whether the given token appears inside a \AngleParenthesis{code} markdown tag in the StackOverflow post. It is worth noting that \AngleParenthesis{code} tags are noisy as users do not always enclose inline code in a \AngleParenthesis{code} tag or sometimes use the tag to highlight non-code texts (details in \S \ref{annotation_scheme}). 
Nevertheless, we find it helpful to include the markdown information as a feature as it improves the performance of our segmentation model.

\end{itemize}

\noindent The inclusion of hand-crafted features is influenced by \citeauthor{wu2018evaluating} \shortcite{wu2018evaluating}, where word-shapes and POS tags were shown to improve the performance of sequence tagging models.

\subsection{Embedding-Level Attention}
\label{sec:han}


For each input word $w_i$ in the input sentence, we have three embeddings: BERT ($w_{i1}$), Code Recognizer ($w_{i2}$), and Entity Segmenter ($w_{i3}$).  We introduce the embedding-level attention ${\alpha}_{it}$ ($t \in \{1, 2, 3\}$), which captures each embedding's contribution towards the meaning of the word, to combine them together. To compute ${\alpha}_{it}$, we pass the input embeddings through a bidirectional GRU and generate their corresponding hidden representations $h_{it} = \overleftrightarrow{GRU}(w_{it})$. These vectors are then passed through a non-linear layer, which outputs $u_{it} = tanh(W_{e}h_{it} + b_{e})$.  We introduce an embedding-level context vector $u_e$, which is randomly initialized and updated during the training process. This context vector is combined  with the hidden embedding representation using a softmax function to extract weight of the embeddings: ${\alpha}_{it} = \frac{exp({u_{it}}^T u_{e})}{{\sum}_t exp({u_{it}}^T u_{e})}$. Finally, we create the word vector by a weighted sum of all the information from different embeddings as $word_{i} = {\sum}_t {\alpha}_{it}h_{it}$. The aggregated word vector $word_{i}$ is then fed into a linear-CRF layer, which predicts the entity category for each word based the BIO tagging schema.

\subsection{Implementation Details}
\label{sec:implementation}

We use PyTorch framework to implement our proposed SoftNER model and its two auxiliary components, namely code recognition and entity segmentation. The input to the SoftNER model include 850-dimensional vectors extracted from both the code recognizer and the entity segmenter. 

We pre-trained BERT\textsubscript{base}, ELMo and GloVe vectors on 152 million sentences from the StackOverflow, excluding sentences from the 1,237 posts in our annotated corpus. The pre-training of the 768-dimensional BERT\textsubscript{base} model with 64,000 WordPiece vocabulary took 7 days on a Google TPU. The pre-training of 1024-dimensional ELMo vectors took 46 days on 3 NVIDIA Titan X Pascal GPUs. The pre-training of 300-dimensional GloVe embeddings \cite{pennington2014glove} with a frequency cut-off of 5 took 8 hours on a server with 32 CPU cores and 386 GB memory.  

We train the SoftNER model and the two auxiliary models separately. Our segmentation model follows the simple BERT fine-tuning architecture except for the input, where BERT embeddings are concatenated with 100-dimensional code markdown and 10-dimensional word frequency features.  We set the number of bins $k$ to 10 for Gaussian vectorization. Our code recognition model is a feedforward network with two hidden layers and a single output node with sigmoid activation.

\section{Evaluation}
\label{sec:experiments}

\begin{table}[!t]
\footnotesize
\centering
\addtolength{\tabcolsep}{-0.05 in} 
\begin{tabular}{l|p{.28 in} p{.28 in} p{.28 in}}

\hline
&  \textbf{ \hspace{.02 in} P} &  \textbf{\hspace{.02 in} R}    & \textbf{\hspace{.02 in} F$_{1}$} 
\\ \hline 

\multicolumn{4}{l}{\textit{\textbf{Test set}}}   \\ \hline 

Feature-based CRF       & 71.77 & 39.70 & 51.12 \\ 
BiLSTM-CRF ({\scriptsize{ELMoVerflow}})& 73.03 & 64.82 & 68.68 \\
Attentive BiLSTM-CRF ({\scriptsize{ELMoVerflow}})   & \underline{78.22} & \underline{78.59} & \underline{78.41}\\ 
Fine-tuned BERT & 77.02 & 45.92   & 57.54 \\ 
Fine-tuned BERTOverflow &  68.77 &  67.47 & 68.12 \\ 
SoftNER ({\scriptsize{BERTOverflow}}) &  \textbf{78.42} & \textbf{79.79} & \textbf{79.10} \\
\hline

\multicolumn{4}{l}{\textit{\textbf{Dev set}}}   \\ \hline 

Feature-based CRF       & 66.85 & 46.19 & 54.64 \\
BiLSTM-CRF ({\scriptsize{ELMoVerflow}})   & 74.44 & 68.71 & 71.46 \\ 
Attentive BiLSTM-CRF ({\scriptsize{ELMoVerflow}})   & \underline{79.43} &  \underline{80.00} &  \underline{79.72} \\ 
Fine-tuned BERT & 79.57 & 46.42   & 58.64 \\ 
Fine-tuned BERTOverflow &  72.11 & 70.51 & 71.30 \\
SoftNER  ({\scriptsize{BERTOverflow}}) &  \textbf{78.81} & \textbf{81.72} & \textbf{80.24} \\
\hline
\end{tabular}

\caption{Evaluation on the \textit{dev} and \textit{test} sets of the StackOverflow NER corpus. Our SoftNER model outperforms the existing approaches.}
\label{tab:main_results}

\end{table}

In this section, we show that our SoftNER model outperforms all the previous NER approaches on the StackOverflow and GitHub data. We also discuss the factors pivotal to the performance of our model, namely pre-trained in-domain BERT embeddings and two domain-specific auxiliary tasks.

\subsection{Data}
\label{subsec:data}
 We train and evaluate our SoftNER model on the StackOverflow NER corpus of 9,352 train, 2,942 development and 3,115 test sentences we constructed in \S \ref{sec:dataset_and_annotation}.   We use the same data for our segmentation model but replace  all the entity tags with an \textsc{i-entity} tag. For the code recognition model, we created a new lexicon of 6,000 unique tokens randomly sampled from the training set of the StackOverflow NER corpus. Each token was labelled independently without context as \textsc{code}, \textsc{ambiguous} or \textsc{non-code} by two annotators with computer science background. The inter-annotator agreement was 0.89, measured by Cohen’s Kappa. After discarding disagreements, we divided the remaining 5,312 tokens into 4,312 train and 1,000 test instances.  Then, we merged \textsc{ambiguous} and \textsc{non-code} categories to facilitate binary classification. We name this dataset of 5312 individual tokens as \textsc{SOLexicon}.

\subsection{Baselines}

\label{sec:baselines}

We compare our model with the following baseline and state-of-the-art approaches:

\begin{itemize}

    \item A \textbf{Feature-based Linear CRF} model which uses the standard orthographic, context and gazetteer features, along with the code markdown tags and handcrafted regular expressions to recognize code entities (details in Appendix \ref{sec:crf_baseline}).
    
    \item A \textbf{BiLSTM-CRF} model with in-domain ELMo embeddings (\textbf{ELMoVerflow}; details in \S \ref{sec:implementation}). This architecture is used as the state-of-the-art baseline named-entity recognition models in various  domains \cite{Lample2016,kulkarni2018annotated,dai2019using}.
    
    \item  An \textbf{Attentive BiLSTM-CRF} model with in-domain ELMo embeddings as well as domain-specific embeddings from the code recognizer and the entity segmenter. This model combines these three word embeddings using an attention network and then utilizes a BiLSTM-CRF layer to predict the entity type of each input word  (details in Appendix \ref{sec:elmo_attention_baseline}).

    \item A \textbf{Fine-tuned out-of-domain BERT} model where we fine-tune the original BERT$_{base}$ cased checkpoint\footnote{\url{https://github.com/google-research/BERT}} on our annotated corpus.
    
    \item A \textbf{Fine-tuned in-domain BERT} model where we fine-tune the in-domain pre-trained BERT$_{base}$ (\textbf{BERTOverflow}; details in \S \ref{sec:implementation}) cased checkpoint\footnote{\url{https://github.com/lanwuwei/BERTOverflow/}} on our annotated corpus.

\end{itemize}

\subsection{Results}

Table \ref{tab:main_results} shows the precision (\textbf{P}), recall (\textbf{R}) and \textbf{F$_1$} score comparison of different models evaluated on the StackOverflow NER corpus. Our SoftNER model outperforms the existing NER approaches in all the three metrics. Fine-tuning over in-domain trained BERT (BERTOverflow), in particular, improves F$_1$ score by more than 10 points in comparison to using the original BERT. 

\subsection{In-domain vs. out-of-domain Word Embeddings}


Table \ref{tab:bilstm_domain} shows the performance comparison between in-domain and out-of-domain word embeddings. We consider off-the-shelf BERT \cite{devlin2018bert},  ELMo \cite{Peters:2018} and GloVe \cite{pennington2014glove} vectors trained on newswire and web texts as out-of-domain embeddings. When using the BiLSTM-CRF model \cite{Lample2016,kulkarni2018annotated,dai2019using}, we observe a large increase of 13.64 F$_{1}$ score when employing in-domain ELMo (ELMoVerflow) representations over in-domain GloVe (GloVeOverflow), and an increase of 15.71 F$_{1}$ score over out-of-domain ELMo. We found that fine-tuning out-of-domain BERT \cite{devlin2018bert} outperforms the out-of-domain ELMo (Table \ref{tab:bilstm_domain}), although it underperforms in-domain ELMo (ELMoVerflow)  by 12.92 F$_1$ score and in-domain BERt (BERTOverflow) by 12.76 F$_1$ score (Table \ref{tab:main_results}). Similarly, in-domain ELMo outperforms the out-of-domain fine-tuned BERT by 10.67 F$_1$ score on Github data (Table \ref{tab:gh_results}; more details in \S \ref{sec:domain_adaptation}). 

It is  worth noting that, the performance improvements from contextual word embeddings are more pronounced on our software domain than on newswire and biomedical domains. Original ELMo and BERT outperform GloVe by 2.06 and 2.12 points in F$_1$ respectively on CoNLL 2003 NER task of newswire data \cite{Peters:2018, devlin2018bert}. For biomedical domain, in-domain ELMo outperforms out-of-domain ELMo by only 1.33 points in F$_1$ on the BC2GM dataset \cite{sheikhshabbafghi-etal-2018-domain}.


We hypothesized that the performance gains from the in-domain contextual embeddings are largely aided by the model's ability to handle ambiguous and unseen tokens. The increase in performance is especially notable (41\% $\xrightarrow{}$ 70\% accuracy) for unseen tokens, which constitute 38\% of the tokens inside gold entity spans in our dataset. This experiment also demonstrates that our annotated NER corpus provides an attractive test-bed for measuring the adaptability of different contextual word representations.


\begin{table}[!t]
\footnotesize
\centering
\begin{tabular}{l|p{.25 in} p{.25 in} p{.25 in}}
\hline
&  \textbf{ \hspace{.02 in} P} &  \textbf{\hspace{.02 in} R}    & \textbf{\hspace{.03 in} F$_{1}$} 
\\ \hline 
\multicolumn{4}{l}{\textit{\textbf{out-of-domain Word Embeddings}}}   \\ \hline 
GloVe (newswire+Wiki+Web) & 61.71 & 49.08 & 54.67 \\
ELMo (newswire+Wiki) & 67.66 & 47.41 & 55.75 \\
Fine-tuned BERT (book+Wiki) & 45.92 & 77.02 & 57.54
\\ \hline
\multicolumn{4}{l}{\textit{\textbf{In-Domain Word Embeddings}}}   \\ \hline 
GloVeOverflow  & 66.28 & 51.28 & 57.82\\
ELMoVerflow & \textbf{74.44} & 68.71 & \textbf{71.46} \\
Fine-tuned BERTOverflow  &  72.11 & \textbf{70.51} & 71.30 \\ 
\hline
\end{tabular}
\caption{Performance of fine-tuned BERT model, BiLSTM-CRF model with GloVe and ELMo embeddings on the {\em dev} set of our StackOverflow NER corpus. Contextualized word representations show a clear benefit when trained on the in-domain StackOverflow data.}
\label{tab:bilstm_domain}
\end{table}



\subsection{Evaluation of Auxiliary Systems}
\label{sec:evaluation_auxiliary}

The domain-specific vectors produced by the \textbf{Code Recognizer} and the \textbf{Entity Segmenter} are also crucial for the overall performance of our SoftNER model. Table \ref{tab:ablation_study} shows an ablation study. Removing code recognizer vectors and entity segmenter vectors results in a drop of 2.19 and 3.69 in F$_{1}$ scores respectively. If we replace embedding-level attention with a simple concatenation of embeddings, the performance also drop by 2.81 F$_1$. In addition, we evaluate the effectiveness of our two domain-specific auxiliary systems on their respective tasks.

\begin{table}[!t] 
\centering
\footnotesize
\begin{tabular}{l|c c c}
\hline
&  \textbf{P} &  \textbf{R}    & \textbf{F$_{1}$}     \\ 
\hline
 SoftNER & \textbf{78.81} & \textbf{81.72} & \textbf{80.24} \\
\;\; -- Embedding Attention & 75.83 & 79.09 & 77.43\\
\;\; -- Code Recognizer & 78.76 & 77.35 & 78.05 \\
\;\; -- Entity Segmenter & 77.82 & 75.32 & 76.55 \\\hline
\end{tabular}

\caption{Ablation study of SoftNER on the {\em dev} set of StackOverflow NER corpus.}
\label{tab:ablation_study}


\end{table}

\begin{table}[t!]
\centering
\small
\begin{tabular}{l|c c c}
\hline
 & \textbf{P} & \textbf{R}  & \textbf{F$_{1}$} \\
 \hline
Token Frequency & 33.33 & 2.25 & 4.22 \\
Most Frequent Label & \textbf{82.21} & 58.59 & 68.42 \\

Our Code Recognition Model  & \textbf{78.43} & 83.33 & \textbf{80.80}\\
$\;\;$ -- Character ngram LMs &  64.13 & \textbf{84.51} & 72.90\\
$\;\;$ -- Word ngram LMs & 67.98 & 72.96 & 70.38\\
$\;\;$ -- FastText Embeddings  & 76.12 & 81.69 & 78.81\\
\hline
\end{tabular}

\caption{Evaluation results and feature ablation of our code recognition model on \textsc{SOLexicon} {\em test} set of 1000 manually labeled unique tokens, which are sampled from the {\em train} set of StackOverflow NER corpus.}
\label{tab:code_token_results}

\end{table}

\paragraph{Code Recognition:} Table \ref{tab:code_token_results} compares the performance of our code recognition model with other baselines on the \textsc{SLexicon} {\em test} set (\S \ref{subsec:data}), which consists of 1,000 random words from the {\em train} set of StackOverflow NER corpus classified as either a code or a non-code token. The baselines include: (i) a Most Frequent Label baseline, which assigns the most frequent label according to the human annotation in \textsc{SOLexicon} {\em train} set; and (ii) a frequency baseline, which learns a threshold over token frequency in the {\em train} set of StackOverflow NER corpus using a decision tree classifier. Our model outperforms both baselines in terms of F$_1$ score. Although the most frequent label baseline achieves better precision than our model, it performs poorly on unseen tokens resulting in a large drop in recall and F$_1$ score.  The ablation experiments show that the FastText word embeddings along with the character and word-level features are crucial for the code recognition model.

 \paragraph{Entity Segmentation:}  Table \ref{tab:segmenter_results} shows the performance of our segmentation model on the {\em dev} set of our StackOverflow corpus, where the entity tags are replaced by an \textsc{I-Entity} tag. Our model achieves an F$_1$ score of 88.09 and with 90.41\% precision and 85.89\% recall. Incorporating word frequency and code markdown feature increases the F$_1$ score by 1.57 and 2.66 points respectively. The low 10.5 F$_1$ score of Stanford NER tagger \cite{manning-EtAl:2014:Stanford_CoreNLP}, which is trained on newswire text, demonstrates the importance of domain-specific tools for the software engineering domain.

\begin{table}[t!]
\centering
\footnotesize
\begin{tabular}{p{1.58 in} |p{.25 in} p{.25 in} p{.25 in}}
\hline
 &  \textbf{ \hspace{.02 in} P} &  \textbf{\hspace{.02 in} R}    & \textbf{\hspace{.03 in} F$_{1}$} \\
 \hline
Stanford NER Tagger & 63.02 & 5.74 & 10.52 \\
Our Entity Segmentation Model & \textbf{90.41} &  \textbf{85.89} &  \textbf{88.09}\\
$\;\;$ -- Word Frequency & 88.32 & 84.79 & 86.52 \\
$\;\;$ -- Code Markdown  & 86.23 & 84.64 & 85.43 \\
\hline
\end{tabular}
\caption{Evaluation of our segmentation model on the \textit{dev} set of the StackOverflow NER corpus.}
\label{tab:segmenter_results}

\end{table}

\subsection{Error Analysis}

Based on our manual inspection, the incorrect predictions made by NER systems on StackOverflow data can be largely classified into the following two categories (see examples in Table \ref{tab:err_example}): 

\begin{itemize}[noitemsep]
    \item \textbf{Segmentation Mismatch} refers to the cases where model predicts the boundary of entities incorrectly. Our SoftNER model reduces such segmentation errors by 89.36\% compared to the fine-tuned BERTOverflow baseline.
    \item \textbf{Entity-Type Mismatch} refers to the errors where a code entity (e.g., names of variables) is predicted as a non-code entity (e.g., names of devices), and vice-versa. Our SoftNER model reduces such entity type errors by 13.54\% compared to the fine-tuned BERTOverflow baseline.
\end{itemize}

\begin{figure}[!t]
\includegraphics[width=3.0 in,keepaspectratio]{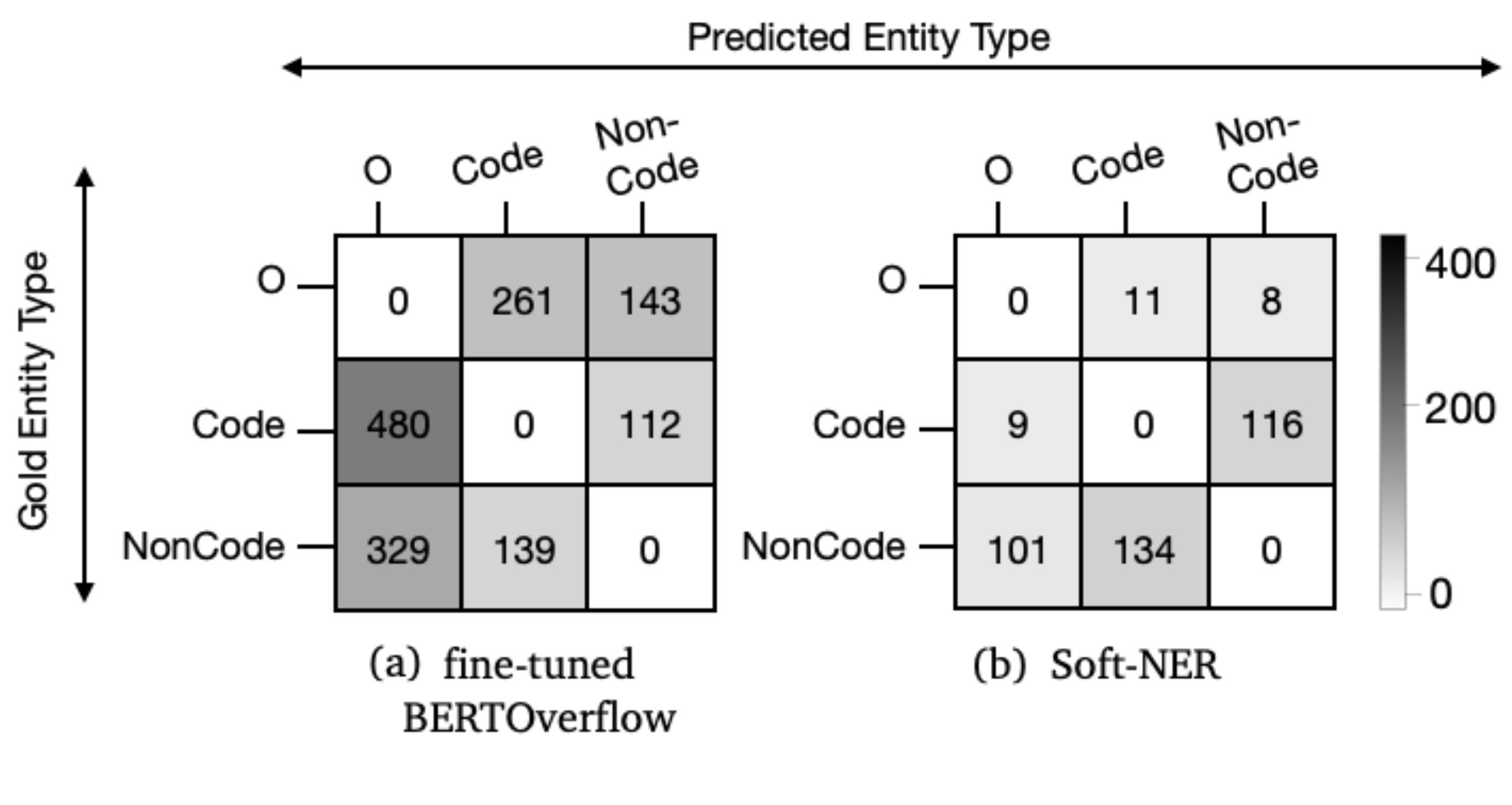}
\caption{Comparison of errors made by the fine-tuned BERTOverflow baseline and our SoftNER model on the {\em dev} set of the StackOverflow NER corpus. In the heatmap,  darker cell color corresponds to higher error counts. Our SoftNER model reduces errors in all the categories.}
\label{fig:error_confusion_matrix}
\end{figure}

\noindent As illustrated in Figure \ref{fig:error_confusion_matrix}, our SoftNER model reduced the errors in both categories by incorporating the auxiliary outputs from segmenter and code recognizer model.

\section{Domain Adaptation to GitHub data}
\label{sec:domain_adaptation}

To  understand the  domain adaptability of our StackOverflow based SoftNER, we evaluate its performance on readme files and issue reports from 143 randomly sampled repositories in the GitHub dump \cite{GHTorrent}. We also trained ELMo embeddings (ELMoGithub) on 4 million sentences from randomly sampled 5,000 GitHub repositories.

\begin{table}[!t]
    \centering
      \begin{tabular}{ p{1.55 cm}|  p{5.0 cm} }
      \hline
      
      \\\\[-4.3\medskipamount]
      { \footnotesize { {\hspace{-.1 in}} Segmentation Mismatch}}
      & 
      \raisebox{-0.8 cm}{\includegraphics[width=0.32\textwidth]{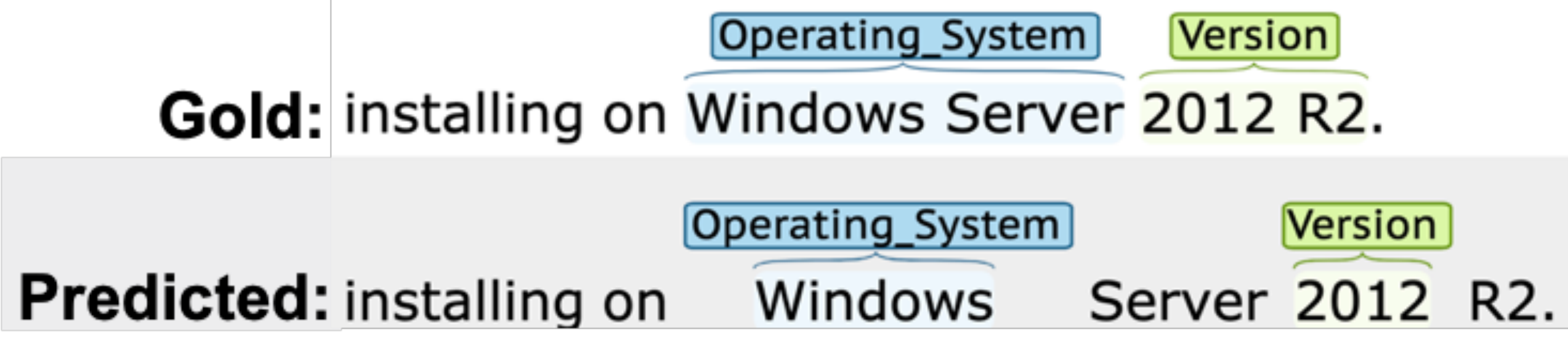}}
      \\ 
      \hline
      { \footnotesize Entity-Type Mismatch}  &
      \raisebox{-0.7 cm}{\includegraphics[width=0.33\textwidth]{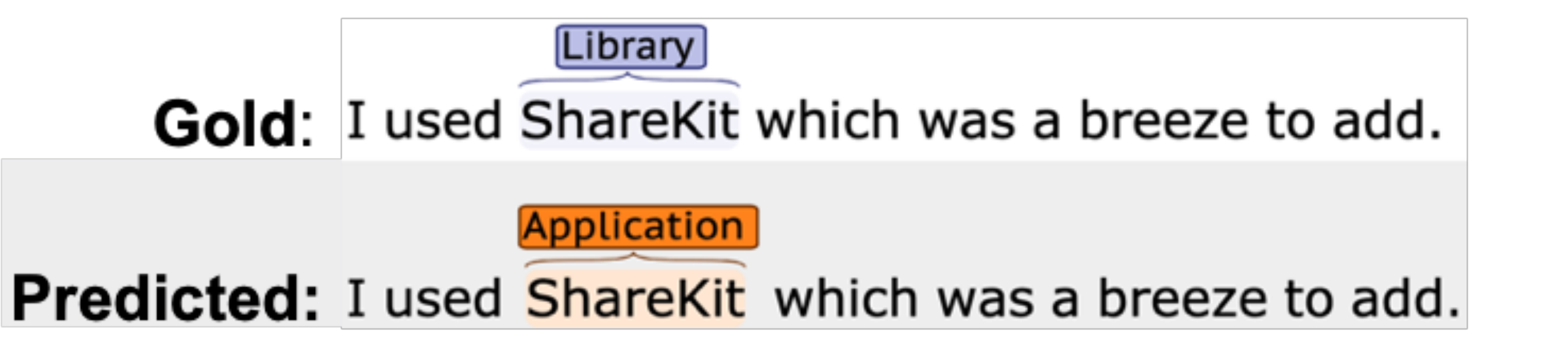}} 
      \\
      \hline 
      \end{tabular}
      \caption{Representative examples of system errors.}
  \label{tab:err_example}
  \end{table}

\begin{table}[!t] 
\centering
\footnotesize
\addtolength{\tabcolsep}{-0.05 in} 
\begin{tabular}{l |p{.28 in} p{.28 in} p{.28 in}}
\hline
 &  \textbf{ \hspace{.02 in} P} &  \textbf{\hspace{.02 in} R}    & \textbf{\hspace{.03 in} F$_{1}$} \\
\hline
 Feature-Based CRF & 43.16 & 35.71 & 39.09 \\
 BiLSTM-CRF ({\scriptsize{ELMoGitHub}}) &  \textbf{64.53} & \textbf{60.96} & \textbf{62.69} \\ 
 Attentive BiLSTM-CRF ({\scriptsize{ELMoVerflow}}) & 62.05 & 59.20 & 60.59 \\ 
 Attentive BiLSTM-CRF ({\scriptsize{ELMoGitHub}})  & \underline{63.29} & \underline{60.89} & \underline{62.07} \\ 
 Fine-tuned out-of-domain BERT  & 56.59 & 48.13  & 52.02 \\
 Fine-tuned BERTOverflow  &  61.71 &  58.75 & 60.19\\
 SoftNER ({\scriptsize{BERTOverflow}}) &  61.92 &  60.26  & 61.08 \\


 \hline
\end{tabular}
 \caption{Evaluation on the GitHub NER dataset of readme files and issue posts. \textit{All the models are trained on our StackOverflow NER corpus.} Our SoftNER model performs close to BiLSTM-CRF model trained on the GitHub ELMo embeddings.}
 \label{tab:gh_results}
\end{table}

Table \ref{tab:gh_results} shows that the performance of our SoftNER model using StackOverflow ELMo embeddings is similar to the top performing BiLSTM-CRF model using GitHub ELMo embeddings with a difference of only 1.61 points in F$_1$. We also did not observe any significant gain after adding the code recognizer and segmenter vectors to the Github ELMo embeddings. We think one likely explanation is that GitHub data contains less code-related tokens when compared to StackOverflow. The percentage of code-related entity tokens is 63.20\% in GitHub and 77.21\% in StackOverflow. Overall, we observe a drop of our SoftNER tagger from 79.10 F$_1$ on StackOverflow (Table \ref{tab:main_results}) to 61.08 F$_1$ on GitHub data (Table \ref{tab:gh_results}) in F$_1$ due to domain mismatch. However, we believe that our NER tagger still achieves sufficient performance to be useful for applications on GitHub.\footnote{As a reference, the state-of-the-art performance for 10-class Twitter NER is 70.69 F$_1$\cite{zhang2018adaptive}.} We leave investigation of semi-supervised learning and other domain adaptation approaches for future work.

\section{Related Work}

The CoNLL 2003 dataset \cite{sang2003introduction} is a widely used benchmark for named entity recognition, which contains annotated newswire text from the Reuters RCV1 corpus.  State-of-the-art approaches on this dataset \cite{baevski-etal-2019-cloze} use a bidirectional LSTM \citep{Lample2016,ma-hovy-2016-end} with conditional random field \cite{colloBERT2011natural} and contextualized word representations \cite{mccann2017learned, Peters:2018,devlin2018bert}.

Named entity recognition has been explored for new domains and languages, such as social media \cite{finin2010annotating,Ritter11,plank2014learning,derczynski2015analysis,limsopatham2016bidirectional,aguilar2017multi}, biomedical texts \cite{collier2004introduction,greenberg2018marginal,kulkarni2018annotated}, multilingual texts \cite{benajiba2008arabic,xie2018neural} and code-switched corpora \cite{aguilar2018named,ball2018part}.  Various methods have been investigated for handling rare entities, for example incorporating external context \cite{long2017world} or approaches that make use of distant supervision \cite{choi-etal-2018-ultra,yang2018distantly,onoe2019learning}.

There has been relatively little prior work on named entity recognition in the software engineering domain.  \citet{ye2016software} annotated 4,646 sentences from StackOverflow with five named entity types (Programming Language, Platform, API, Tool-Library-Framework and Software Standard).  The authors used a traditional feature-based CRF to recognize these entities.  In contrast, we present a much larger annotated corpus consisting of 15,372 sentences labeled with 20 fine-grained entity types. We also develop a novel attention based neural NER model to extract those fine-grained entities.

\section{Conclusion}
 In this work, we investigated the task of named entity recognition in the social computer programming domain.  We developed a new NER corpus of 15,372 sentences from StackOverflow and 6,510 sentences from GitHub annotated with 20 fine-grained named entities. We demonstrate that this new corpus is an ideal benchmark dataset for contextual word representations, as there are many challenging ambiguities that often require long-distance context to resolve. We also proposed a novel attention based model, named SoftNER, that outperforms the state-of-the-art NER models on this dataset.  Furthermore, we investigated the important sub-task of code recognition. Our code recognition model captures additional spelling information beyond then contextual word representations and consistently helps to improve the NER performance.  We believe our corpus, StackOverflow-specific BERT embeddings and named entity tagger  will be useful for various language-and-code tasks, such as code retrieval, software knowledge base extraction and automated question-answering.

\section*{Acknowledgement}
We thank anonymous reviewers for their thoughtful comments. We also thank NVIDIA, Google, and Ohio Supercomputer Center \cite{Oakley2012} for providing GPU/TPU computing resources; Wuwei Lan for kindly helping to train in-domain BERT on StackOverflow data; Sydney Lee, Rita Tong, Lillian Chow, and Raleigh Potluri for help with data annotation. This research is supported in part by the NSF awards IIS-1822754 and IIS-1845670, ODNI and IARPA via the BETTER program contract 19051600004, ARO and DARPA via the SocialSim program contract W911NF-17-C-0095, Criteo Faculty Research Award to Wei Xu, and Amazon Faculty Research Award to Alan Ritter. The views and conclusions contained herein are those of the authors and should not be interpreted as necessarily representing the official policies, either expressed or implied, of NSF, ODNI, IARPA, ARO, DARPA or the U.S. Government. The U.S. Government is authorized to reproduce and distribute reprints for governmental purposes notwithstanding any copyright annotation therein.

\bibliographystyle{acl_natbib}
\bibliography{references}

\clearpage

\appendix

\section{Feature-Based CRF Baseline}
\label{sec:crf_baseline}
We implemented a CRF baseline model using CRFsuite\footnote{{http://www.chokkan.org/software/crfsuite/}} to extract the software entities.  This model uses standard  orthographic, contextual and gazetteer features. It also includes the code markdown tags (\S \ref{sec:segmenter}) and  a set of regular expression features. The regular expressions are developed to recognize specific categories of code-related entities. Feature ablation experiments on this CRF model are presented in Table \ref{tab:feature_crf}. One noticeable distinction from the named entity recognizer in many other domains is that the \textit{contextual features are not as helpful in feature-based CRFs for classifying software entities}. This is because, in the StackOverflow NER corpus a significant number of neighbouring words are shared among different software entities.  As an example, the bigram \textit{`in the'} frequently appears as the left context of the following types: {\sc Application, Class, Function, File Type, UI Element, Library, Data Structure} and {\sc Language}.  

\begin{table}[!h]
\centering
\footnotesize
\begin{tabular}{l|c c c}
\hline
  &  \textbf{P} &  \textbf{R}    & \textbf{F$_{1}$}     \\ 
 \hline
Feature-based CRF &  66.85 & 46.19 & 54.64 \\
$\;\;$ -- Context Features & 68.91 & 43.58 & 53.39\\
$\;\;$ -- Markdown Feature & 70.64 & 40.15 & 51.20\\
$\;\;$ -- Rule and Gazetteer Features & 69.71 & 40.66 & 51.36\\

\hline
\end{tabular}
\caption{Feature based CRF performance with varying input features on {\em dev} data.}
\label{tab:feature_crf}
\end{table}

\section{Attentive BiLSTM CRF with ELMoVerflow }
\label{sec:elmo_attention_baseline}

We propose a baseline Attentive NER model that utilizes a BiLSTM-CRF network to predict the entity type of each word from its weighted representations. The weighted word representations are extracted by a multi-level attention network, similar to \citeauthor{yang-etal-2016-hierarchical}\shortcite{yang-etal-2016-hierarchical}, that combines the contextualized ELMo embeddings with the code recognizer (\S \ref{sec:inline_code_recognition}) and segmenter vector (\S \ref{sec:segmenter_elmo}). These three  input embeddings are merged together in the first attention layer and then their corresponding weights are calculated using the second layer. Although such multi-level attention is not commonly used in NER, we found it empirically helpful for the software domain (see Table \ref{tab:ablation_study_attn_ner}). 

\begin{table}[!t] 
\centering
\footnotesize
\begin{tabular}{l|c c c}
\hline
&  \textbf{P} &  \textbf{R}    & \textbf{F$_{1}$}     \\ 
\hline
Attentive BiLSTM-CRF &  \textbf{79.43} &  \textbf{80.00} &  \textbf{79.72}\\
 \;\; -- Multi-level Attention & 77.68 & 78.08 & 77.88\\
\;\; -- Code Recognizer & 77.18 & 77.76 & 77.47 \\
\;\; -- Entity Segmenter & 74.82 & 75.32 & 75.07 \\\hline
\end{tabular}

\caption{Ablation study of Attentive-NER on the {\em dev} set of StackOverflow NER corpus.}
\label{tab:ablation_study_attn_ner}

\vspace{-7 mm}

\end{table}

\begin{figure*}[tbhp]
    \centering
    \includegraphics[width=6.1 in,keepaspectratio]{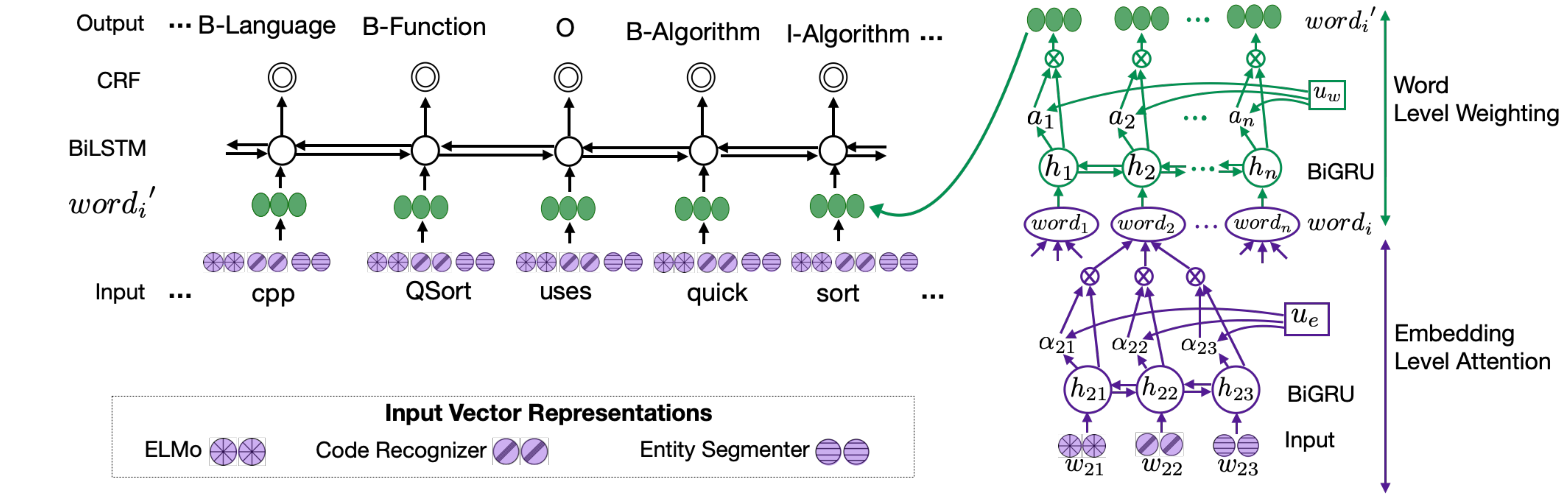}
    \caption{Our Attentive BiLSTM-CRF model. It utilizes an attention network to combine the contextual word embeddings (ELMo) with the domain-specific embeddings (Code Recognizer and Entity Segmenter). The detailed structure of the attention network is depicted on the right.}
    \label{fig:han_model}
    
\end{figure*}

\paragraph{Embedding-Level Attention}  uses three embeddings, ELMo ($w_{i1}$), Code Recognizer ($w_{i2}$), and Entity Segmenter ($w_{i3}$), for each word $w_i$ in the input sentence. The embedding-level attention ${\alpha}_{it}$ ($t \in \{1, 2, 3\}$) to captures each embedding's contribution towards the meaning of the word. To compute ${\alpha}_{it}$, it pass the input embeddings through a bidirectional GRU and generate their corresponding hidden representations $h_{it} = \overleftrightarrow{GRU}(w_{it})$. These vectors are then passed through a non-linear layer, which outputs $u_{it} = tanh(W_{e}h_{it} + b_{e})$.  It uses an embedding-level context vector, $u_e$, which is learned during the training process.  This context vector is combined  with the hidden embedding representation using a softmax function to extract weight of the embeddings, ${\alpha}_{it} = \frac{exp({u_{it}}^T u_{e})}{{\sum}_t exp({u_{it}}^T u_{e})}$. Finally, the word vector is created by a weighted sum of all the information from different embeddings as $word_{i} = {\sum}_t {\alpha}_{it}h_{it}$.

\paragraph{Weighted Word Representation} uses a word-level weighting factor $\alpha_{i}$ to emphasize the importance of each word $w_i$ for the NER task. Similar to the embedding-level attention, it calculates $\alpha_{i}$ from the weighted word vectors $word_{i}$. A bidirectional GRU is used to encode the summarized information from neighbouring words and thus it get $h_{i} = \overleftrightarrow{GRU}(word_{i})$. This is then passed through a hidden layer which outputs $u_{i} = tanh(W_{w}h_{i} + b_{w})$. Then the normalized weight for each word vector is extracted by  ${\alpha}_{i} = \frac{exp({u_{i}}^T u_w)}{{\sum}_t exp({u_{i}}^T u_w)}$, where $u_w$ is another word-level context vector that is learned during training. The final  weighted word representation is computed by $word_{i}' =  {\alpha}_{i}h_{i}$. 

Subsequently, the aggregated word vector $word_{i}'$ is fed into a BiLSTM-CRF network, which predicts the entity category for each word. The complete architecture of the Attentive BiLSTM CRF model is illustrated in Figure \ref{fig:han_model}. Compared to BiLSTM-CRF, our proposed Attentive BiLSTM-CRF demonstrates a 9.7 increase in F$_1$ on the {\em test} set (Table \ref{tab:main_results}) and reduces the segmentation errors and entity type errors by 80.33\% 23.34\% respectively.


\section{Entity Segmentation with ELMoVerflow}\label{sec:segmenter_elmo}

The Attentive-NER tagger utilizes the outputs from an auxiliary segmentation module which consists of a BiLSTM encoder and a CRF decoder. This model concatenates ELMo embeddings with two hand-crafted features- word frequency and code markdown (\S \ref{sec:segmenter}). 

The segmentation model follows the same architecture and training setup as the Attentive-NER model except for the input, where ELMo embeddings are concatenated with 100-dimensional code markdown and 10-dimensional word frequency features. The binary output from this entity segmenter model is later passed as through an embedding layer and  used as one of the auxiliary inputs of the Attentive NER model.

Table \ref{tab:segmenter_results_elmo} shows the performance of this segmentation model with ELMoVerflow on the {\em dev} set. This model achieves an F$_1$ score of 84.3 and an accuracy of 97.4\%. The ablation study in Table \ref{tab:segmenter_results_elmo} depicts the importance of the hand-crafted  frequency and markdown features for this segmenter model by providing an increment of 1.2 and 2.1 points in the F$_1$ score respectively. 

\begin{table}[!h]
\centering
\footnotesize
\begin{tabular}{p{1.58 in} |p{.25 in} p{.25 in} p{.25 in}}
\hline
 &  \textbf{ \hspace{.02 in} P} &  \textbf{\hspace{.02 in} R}    & \textbf{\hspace{.03 in} F$_{1}$} \\
 \hline

Entity Segmentation {\tiny(ELMoVerflow)}  & \textbf{86.80} &  \textbf{81.86} &  \textbf{84.26}\\
$\;\;$ -- Word Frequency & 84.61 & 81.53 & 83.04 \\
$\;\;$ -- Code Markdown  & 82.49 & 81.83 & 82.16\\
\hline
\end{tabular}
\caption{Ablation study of our segmentation model with ELMoVerflow on the \textit{dev} set of the StackOverflow NER corpus.}
\label{tab:segmenter_results_elmo}

\end{table}

\end{document}